\documentclass[letterpaper, 10 pt, conference]{ieeeconf}  

\IEEEoverridecommandlockouts                              

\usepackage{amsmath}
\usepackage{xspace}

\usepackage[utf8]{inputenc} 
\usepackage[T1]{fontenc}    
\usepackage{hyperref}       
\usepackage{url}            
\usepackage{booktabs}       
\usepackage{amsfonts}       
\usepackage{nicefrac}       
\usepackage{microtype}      
\usepackage{graphicx}
\usepackage{amssymb}
\usepackage{wrapfig}
\usepackage{dsfont}
\usepackage[ruled, noend]{algorithm2e}
\usepackage{mathrsfs}
\usepackage{multicol}
\usepackage{threeparttable}
\usepackage{multirow}
\usepackage{comment}
\let\labelindent\relax
\usepackage{enumitem}
\usepackage{bm}
\usepackage{pifont}
\usepackage{threeparttable}
\usepackage{dsfont}
\usepackage{makecell}
\usepackage[dvipsnames, table]{xcolor}
\usepackage{caption}
\usepackage{subcaption}
\usepackage[export]{adjustbox}
\usepackage{tcolorbox}
\usepackage{cite}
\definecolor{darkgreen}{rgb}{0.0, 0.5, 0.0}
\definecolor{navyblue}{rgb}{0.0, 0.0, 0.5}

\title{\LARGE \bf
Ergodic Imitation for Adaptive Exploration around Demonstrations
}

\author{Ziyi Xu$^{1*}$, Cem Bilaloglu$^{2,1*}$, Yiming Li$^{2,1}$, and Sylvain Calinon$^{2,1}$ 
\thanks{*Equal contribution.}%
\thanks{The authors are with the $^{1}$ Ecole Polytechnique F{\'e}d{\'e}rale de Lausanne (EPFL), Switzerland and with the $^{2}$ Idiap Research Institute, Martigny, Switzerland {\tt\small ziyi.xu@epfl.ch, cem.bilaloglu@idiap.ch, yiming.li@idiap.ch, sylvain.calinon@idiap.ch}
}
}

\begin{document}

\maketitle
\thispagestyle{empty}
\pagestyle{empty}

\begin{abstract}
    In robotics, a common challenge in imitation learning is the mismatch between training and deployment conditions, caused, for example, by environmental changes or imperfect observation and control. When a robot follows a nominal trajectory under such mismatch, it may become stuck and fail to complete the task. This calls for adaptive online exploration strategies that remain grounded in demonstrations. To this end, we propose an adaptive ergodic imitation approach that constructs a target distribution from the geometry of the retrieved demonstrations and uses it to generate trajectories that adaptively interpolate between tracking and exploration. Our method extends ergodic control beyond its traditional role in area-coverage and search by incorporating demonstrations into a retrieval-based receding-horizon framework for adaptive imitation. 

\end{abstract}

\section{INTRODUCTION}

\label{sec:intro}
\begin{figure*}[t]
    \centering
    \includegraphics[width=\textwidth]{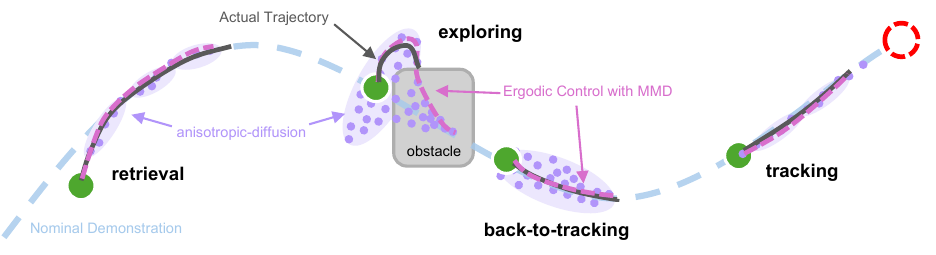}
    \caption{
    Overview of \textsc{Adaptive Ergodic Imitation}. A nominal trajectory induces tracking behavior when execution remains aligned with the demonstration. Under mismatch, the target particle distribution expands and the ergodic planner promotes exploration around the reference. Once the obstacle is bypassed, the score-based kernel contracts the distribution and pulls the agent back toward the demonstrated trajectory.
    }
    \label{fig:overview}
\end{figure*}
Imitation learning (IL) is a practical paradigm for robot programming, where the intended behavior is learned by demonstrations without an explicit reward. In practice, however, the objective is rarely to replicate a demonstration exactly; rather, the agent must adapt the observed behaviors to novel environments that might diverge from the training distribution. Recent critiques of deep generative models in robotics suggest that these architectures often overfit the demonstration data, essentially memorizing specific action sequences rather than learning generalizable policies~\cite{he2025demystifying,Li26ICLR}. Consequently, current IL approaches remain notoriously brittle even under minimal distribution shifts. This necessitates a shift toward imitation paradigms that prioritize situational adaptation over pure trajectory replay.

Although recent adaptive exploration methods allow systems to sample different modes from the dataset, they are largely restricted to discrete transitions~\cite{razmjoo2025ccdp, jin2025sime}. Consequently, they lack the continuous exploration capabilities needed for the subtle, state-space adjustments typical of tasks like robotic assembly. We employ a different perspective that treats demonstrations as references to be tracked under nominal conditions, by using them as informed priors for exploration when environmental shifts render pure tracking insufficient. Instead of considering tracking and exploring as two different objectives, we use the ergodic control methodology to formulate tracking as a special case of exploration. This results in a continuous behavioral spectrum, from rigid imitation to adaptive exploration, governed by the same underlying controller.

Ergodic control synthesizes trajectories whose time-averaged state visitation statistics converge to a target spatial distribution \cite{mathewSpectralMultiscaleCoverage2009}. Traditionally, this framework is employed for exploration \cite{millerErgodicExplorationDistributed2016}, area coverage \cite{bilalogluTactileErgodicCoverage2025}, or search operations \cite{ivicSearchStrategyComplex2020}, using mobile robots and UAVs. More recently, several works have extended ergodic control to the IL setting, applying it to tasks such as robotic insertion \cite{shettyErgodicExplorationUsing2022, sunFastErgodicSearch2024}, cart-pole inversion, and surface cleaning \cite{kalinowskaErgodicImitationLearning2021a}. In this imitation context, demonstrations are leveraged to construct a static target distribution that the ergodic controller used for task reproduction. A primary advantage of ergodic imitation is its reliance on statistical state-visitation rather than strict temporal sequencing; this allows the robot to synthesize successful behaviors without being tethered to the demonstrator's specific timeline~\cite{kalinowskaErgodicImitationLearning2021a}. While this flexibility is beneficial for tasks like cleaning or insertion, temporal dependency remains critical for most manipulation tasks. To address this, we propose an adaptive formulation that strikes a balance between these two paradigms. Our approach reproduces the temporal characteristics of the demonstration when the environment permits, yet autonomously transitions to ergodic exploration if the robot becomes "stuck" and can not track the reference motion. 



In this work, we introduce an adaptive ergodic imitation approach that brings the target distribution-driven exploration mechanism of ergodic control into imitation learning for runtime adaptation. Our work positions ergodic control as a promising direction for generalization in imitation learning, not only as a tool for search or coverage, but as a principled mechanism for adaptive execution around demonstrations. Our contributions are as follows:
\begin{itemize}[leftmargin=0.2in]
\item a unified method for a continuous spectrum of tracking and exploration behaviors using adaptive ergodic imitation;
\item task progress estimation via demonstration to adaptively modify target distributions;
\item geometry-guided anisotropic diffusion for synthesizing target distributions inducing tracking and exploration;
\item using Maximum Mean Discrepancy (MMD) ergodic metric~\cite{hughes2025ergodic} in a retrieval-based imitation learning framework.
\end{itemize}

\vspace{3mm}

\section{METHODOLOGY}
We consider a dataset $\mathcal{D}=\left\{\Gamma^{(i)}\right\}_{i=1}^N$ of $N$ expert demonstrations, where each demonstration is a state trajectory $\Gamma^{(i)}=\left\{\boldsymbol{x}_t^{(i)}\right\}_{t=0}^{T_i}$. Since the proposed framework utilizes an ergodic controller to derive control laws based on target distributions, the expert actions $\boldsymbol{u}_t^{(i)}$ are omitted, and we focus exclusively on the state trajectories.  We define the aggregate state set $\mathcal{P}=\bigcup_{i, t}\left\{\boldsymbol{x}_t^{(i)}\right\}$ as the flattened collection of all expert states.

\subsection{Background}
A dynamical system's time-averaged state trajectory  defines an empirical distribution, or coverage, as $p(t, \bm{x})=\frac{1}{t} \int_0^t \delta(\bm{x}-\bm{x}(\tau)) d \tau$, where $\delta(\cdot)$ is the Dirac delta function. The system is ergodic with respect to a target distribution $q(\bm{x}) \in \mathcal{S}(\mathcal{X})$ if its coverage converges weakly to the target:
\begin{equation}
    \lim _{T \rightarrow \infty} \mathbb{E}_{\bm{x} \sim p_T}[\phi(\bm{x})]=\mathbb{E}_{\bm{x} \sim q}[\phi(\bm{x})], \quad \forall \phi \in \mathcal{C}(\mathcal{X}).
\end{equation}
The objective of ergodic control is to synthesize control commands that ensure the system's long-term coverage matches the desired distribution $q(\bm{x})$, typically by minimizing a discrepancy measure between $p$ and $q$. MMD provides a discrepancy measure for ergodicity~\cite{hughes2025ergodic}, particularly effective when the target distribution is known only through a set of discrete samples $\left\{\bm{q}_i\right\}_{i=1}^N \subset \mathcal{X}$. For a trajectory represented by  discrete points $\bm{x}_t$, the squared MMD between the trajectory distribution $p_{\bm{x}}$ and the target distribution $q$ is approximated as:
\begin{align}
\label{eq:mmd}
\overline{\mathrm{MMD}}&_k^2(p, q)
=
\frac{1}{T^2}\sum_{t=0}^{T-1}\sum_{t'=0}^{T-1}
k(\bm{x}_t, \bm{x}_{t'}) \notag \\
&
-\frac{2}{TN}\sum_{t=0}^{T-1}\sum_{i=1}^{N}
k(\bm{x}_t, \bm{q}_i)
+z(q),
\end{align}
where $z(q) = \frac{1}{N^2}\sum_{i=1}^N \sum_{j=1}^N k(\bm{q}_i, \bm{q}_j)$ is a constant term that depends solely on the target distribution.

\subsection{Phase Retrieval}
\label{sec:phase_retrieval}
At each re-planning interval, the system queries the dataset $\mathcal{D}$ using the current robot state $\boldsymbol{x}_q=\boldsymbol{x}(t)$ to identify the most relevant expert context. We define the projected state $\boldsymbol{x}^{\prime}$ as the element in $\mathcal{P}$ that minimizes the distance to the query state:
\begin{equation}
\label{eq:projection}
\boldsymbol{x}^{\prime} = \underset{\boldsymbol{x}_k^{(i)} \in \mathcal{P}}{\arg \min } \|\boldsymbol{x}_q - \boldsymbol{x}_k^{(i)}\|_2.
\end{equation}

Each $\bm{x}^{\prime}$ is associated with a specific temporal index $t^{\prime}$ from its source trajectory $\Gamma^{(i)}$, representing the "expert phase".

To evaluate progress relative to the demonstration, we introduce a \emph{phase error} $e(t)= \tau(t)-t^{\prime}$, where $\tau(t)$ is a virtual reference clock. Unlike the continuous execution time $t$, the reference clock $\tau(t)$ adapts to the robot's performance. The update logic for the reference clock and a stagnation counter $c$ is governed by a threshold $\epsilon$ :

\begin{itemize}
\item Progressing $\rightarrow$ Tracking: If $e(t) \leq \epsilon$, the robot is successfully tracking the demonstration. We set $c=0$ and allow the reference clock to advance ( $\dot{\tau}=1$ ).
\item Stagnating $\rightarrow$ Exploration: If $e(t)>\epsilon$, the robot is lagging behind the expert context. We fix the reference clock $(\dot{\tau}=0)$ and increment the stagnation counter $c$ to signal the exploration. 
\end{itemize}
This stagnation signal $c$ provides a heuristic measure of task progress and determines whether to track or explore around demonstrations.

\subsection{Distribution Generation}

We use anisotropic diffusion to generate a particle-based distribution $\left\{\bm{q}_j\right\}_{j=1}^N \subset \mathcal{X}^{(i)}$ along the nominal trajectory $\Gamma^{(i)}$ that unifies \emph{tracking} and \emph{exploration} within a single stochastic differential equation (SDE):
\begin{equation}
\begin{aligned}
\mathrm{d}\bm q_j
&=
\Big[
\underbrace{\kappa(\theta)\bigl(\bm x^{(i)*}(\bm q_j)-\bm q_j\bigr)}_{\text{nominal attraction}}
+
\underbrace{\alpha(\theta)\,\nabla_{\bm q_j} \log p_t(\bm q_j)}_{\text{heat-kernel score}}
\Big]\mathrm{d}t
\\
&\quad+
\underbrace{\sqrt{2\,\mathbf \Sigma(\theta,\bm q_j)}}_{\text{anisotropic diffusion}}
\,\mathrm{d}\mathbf W_t ,
\end{aligned}
\label{eq:sde}
\end{equation}
where $\bm x^{(i)*}(\bm q_j)$ denotes the projection of particle $\mathbf q_j$ onto $\Gamma^{(i)}$, \(\mathbf W_t\) is a standard Wiener process, and \(\theta\in[0,1]\) is a temperature-like progress variable that modulates the balance between tracking and exploration. For small \(\theta\), attraction and score-based tracking dominate; as \(\theta\) increases, these terms weaken and diffusion becomes prominent.

\paragraph{Curve attraction}
The first drift term pulls each particle toward its nearest point on $\Gamma^{(i)}$. Temperature-dependent coefficient $\kappa(\theta)$ is large at \textit{tracking} and weakens at \textit{exploring}.

\paragraph{Heat-kernel score}
The second drift term is the score function of a kernel density estimate defined over $\Gamma^{(i)}$, $p_t(\bm q_j)=\frac{1}{T_i}\sum_{t=0}^{T_i} k_t(\bm q_j,\bm x_t^{(i)})$,
where \(k_t\) is a heat kernel and \(\bm x_t^{(i)}\) are samples along $\Gamma^{(i)}$. The corresponding score hence takes the form of a kernel-weighted average of the individual kernel scores:
\begin{equation}
\nabla_{\bm q_j} \log p_t(\bm q_j)
=
\frac{\sum_{T_i} k_t(\bm q_j,\bm x_t^{(i)})\,\nabla_{\bm q_j}\log k_t(\bm q_j,\bm x_t^{(i)})}
{\sum_{T_i} k_t(\bm q_j,\bm x_t^{(i)})} .
\end{equation}
This term biases particles toward regions of high reference density, helping them stay close to the reference distribution during the tracking phase.

\paragraph{Anisotropic diffusion}
To promote goal-aware exploration around the reference trajectory, we design the diffusion to be anisotropic using the geometry of the demonstrations: noise is larger in directions normal to the trajectory than along its tangent, and larger near the beginning of the trajectory than near its end. Let \({\hat{\bm t}}(\bm x^{(i)*}(\bm q_j))\) denote the local unit tangent at the reference trajectory projection of particle $\bm q_j$. The diffusion term is split as:
\begin{equation}
\begin{aligned}
\sqrt{2\,\mathbf \Sigma(\theta, \bm q_j)}\,\mathrm{d}W_t
&=
\sqrt{2D_{\parallel}(\theta)}\,
\bigl({\hat{\bm t}}{\hat{\bm t}}^\top\bigr)\,\mathrm{d} \mathbf W_t \\
&\quad+
\sqrt{2D_{\perp}(\theta)}\,
\bigl(\mathbf  I-{\hat{\bm t}}{\hat{\bm t}}^\top\bigr)\,\mathrm{d}\mathbf W_t ,
\end{aligned}
\end{equation}
where \(D_{\parallel}\) and \(D_{\perp}\) are the tangential and orthogonal diffusion coefficients, respectively. Choosing \(D_{\perp}\gg D_{\parallel}\) promotes exploration primarily in directions normal to the trajectory while preserving coherence along it.

Particle diffusion is further bounded by a Laplacian envelope defined along \(\Gamma^{(i)}\), parameterized by arc length \(s\), with \(A\) controlling its amplitude and \(b\) its spatial decay:
\begin{equation}
E(s)
=
\frac{A}{2b}
\exp\!\left(
-\frac{| s|}{b}
\right).
\label{eq:envelope}
\end{equation}
As the phase error $e(t)$ and stagnation signal \(c\) increases, the envelope broadens, allowing exploration over a larger neighborhood of the reference trajectory and eventually approaching the uniform case corresponding to pure exploration.

\subsection{Coverage-Aware Ergodic Control with MMD}

Since we use a sample-based representation of the target distribution, we adopt the MMD ergodic metric introduced by Hughes~\emph{et al.}. We implement a receding horizon controller and include the last ten trajectories from previous planning steps in the MMD objective (Eq.~\eqref{eq:mmd}). This allows previously covered regions to be reflected in the objective, encouraging each new plan to complement rather than retrace past coverage.

MMD and our method extend naturally to \(SE(3)\) and to any other curved space where we can define a geometry-aware kernel such as the heat kernel~\cite{borovitskiyMateRnGaussian}, since we can define the MMD objective. In \(SE(3)\) we can exploit the product structure \(SE(3)\cong SO(3)\times \mathbb{R}^3\), such that the resulting kernel captures rotational and translational components in a unified representation.

\section{RESULTS}
\label{sec:results}

\begin{figure}[t]
    \centering
    \includegraphics[width=\columnwidth]{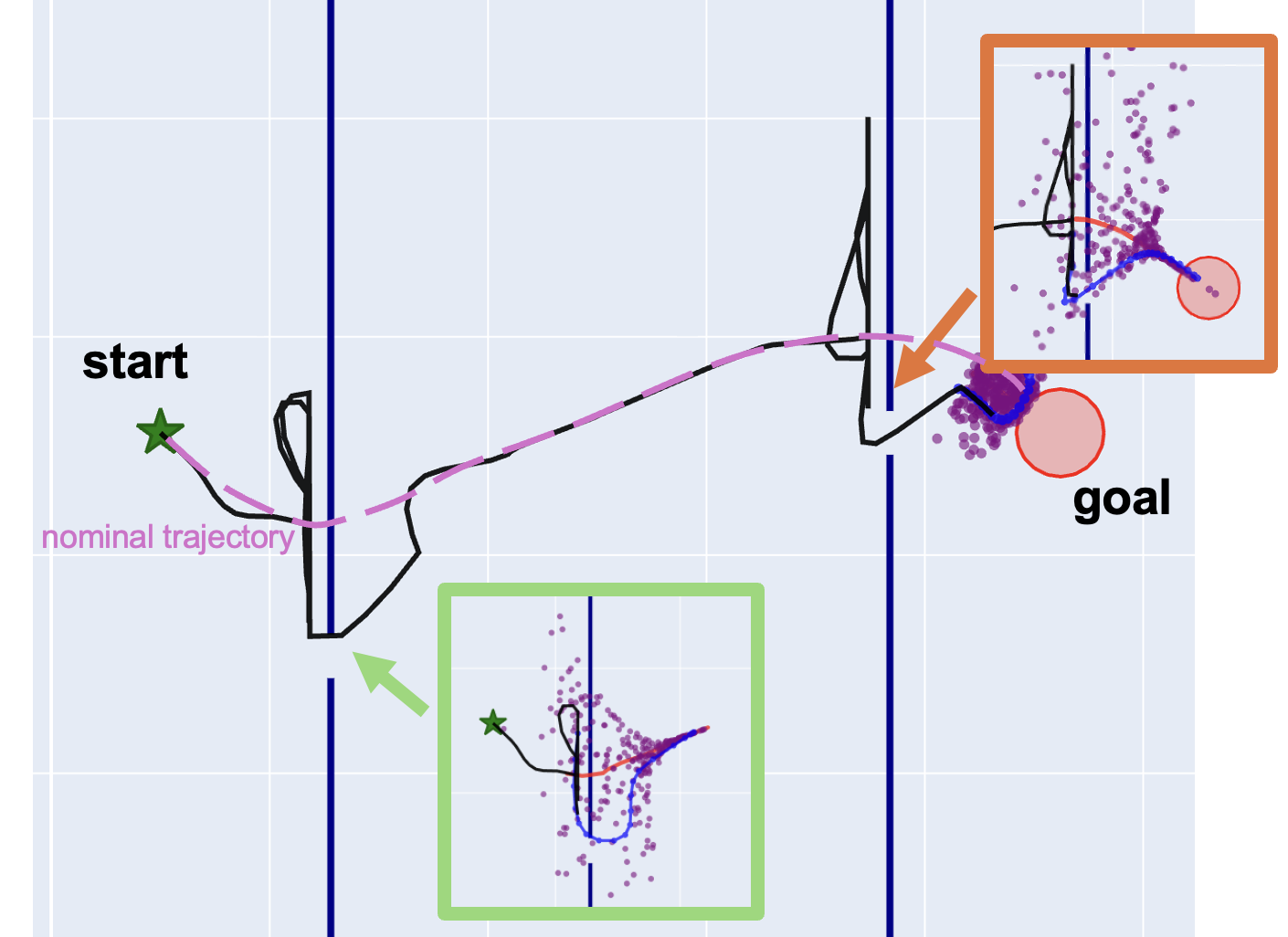}
    \caption{
    Adaptive exploration in the maze environment.
    }
    \label{fig:maze}
\end{figure}

We evaluated our method in a 2D navigation environment with narrow vertical gaps and a cluttered goal region, shown in Fig.~\ref{fig:maze}. Successful demonstrations are first collected in a nominal layout, after which the gap locations are shifted at test time to induce deployment mismatch. Under this perturbation, the agent must adaptively explore around the demonstrated behavior rather than simply replay the nominal trajectory. When blocked by a wall, the agent detects a task-progress mismatch via the accumulated phase error. This discrepancy broadens the target distribution with anisotropic diffusion, resulting in exploration through the ergodic controller.

\begin{figure}[t]
    \centering
    \includegraphics[width=\columnwidth]{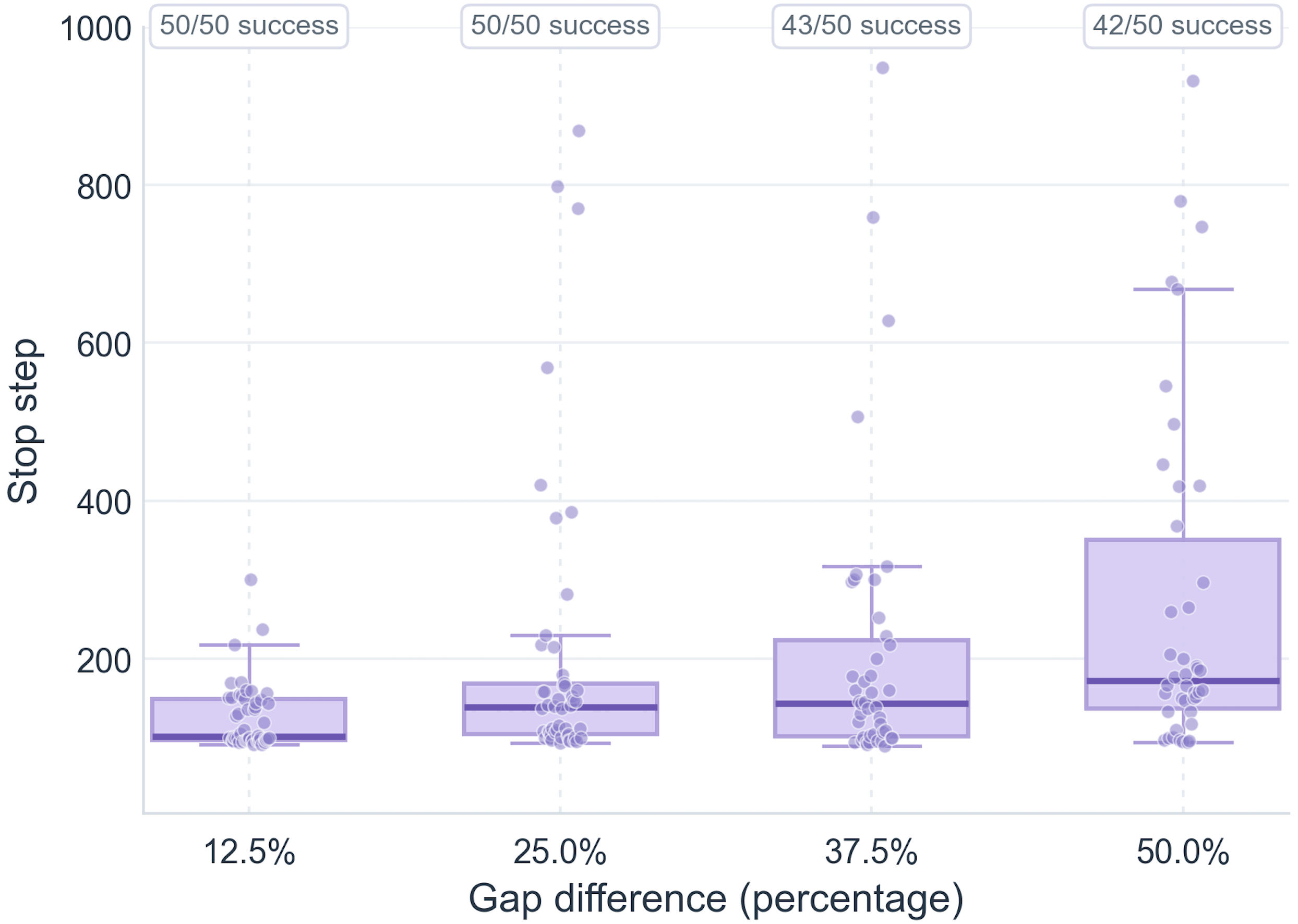}
    \caption{
    Quantitative maze results under gate-location perturbations sampled around the nominal layout. 
    }
    \label{fig:maze_quantitative}
\end{figure}

We additionally tested 50 gate location offsets with respect to the nominal gate position around different gap differences using a Gaussian, shown in Fig.~\ref{fig:maze_quantitative}. \textit{Success} is defined as reaching the goal within 1000 steps, and we used the same nominal trajectory. In such obstacle blocking case, retrieval or generative-based methods would have 0 success rate due to the out-of-distribution gap position, whereas for our method we would always eventually find a solution.



\bibliographystyle{IEEEtran}
\bibliography{reference}



\end{document}